\documentclass{article}

\usepackage{times}
\usepackage{graphicx} % more modern
\usepackage{subfigure} 

\usepackage{natbib}

\usepackage{algorithm}
\usepackage{algorithmic}

\usepackage{hyperref}

\usepackage[accepted]{icml2015}

\icmltitlerunning{A Neural Conversational Model}

\begin{document} 

\twocolumn[
\icmltitle{A Neural Conversational Model}

% It is OKAY to include author information, even for blind
% submissions: the style file will automatically remove it for you
% unless you've provided the [accepted] option to the icml2015
% package.
\icmlauthor{Oriol Vinyals}{vinyals@google.com}
\icmladdress{Google}
\icmlauthor{Quoc V. Le}{qvl@google.com}
\icmladdress{Google}

% You may provide any keywords that you 
% find helpful for describing your paper; these are used to populate 
% the "keywords" metadata in the PDF but will not be shown in the document
\icmlkeywords{neural networks, dialog systems, chatbots}

\vskip 0.3in
]

\begin{abstract}

Conversational modeling is an important task in natural language
understanding and machine intelligence.  Although previous approaches
exist, they are often restricted to specific domains (e.g., booking an
airline ticket) and require hand-crafted rules. In this paper, we
present a simple approach for this task which uses the recently
proposed sequence to sequence framework. Our model converses by
predicting the next sentence given the previous sentence or sentences
in a conversation. The strength of our model is that it can be trained
end-to-end and thus requires much fewer hand-crafted rules. We find
that this straightforward model can generate simple conversations
given a large conversational training dataset. Our preliminary results
suggest that, despite optimizing the wrong objective function, the
model is able to converse well. It is able extract knowledge from both
a domain specific dataset, and from a large, noisy, and general domain
dataset of movie subtitles. On a domain-specific IT helpdesk dataset,
the model can find a solution to a technical problem via
conversations.  On a noisy open-domain movie transcript dataset, the
model can perform simple forms of common sense reasoning. As expected,
we also find that the lack of consistency is a common failure mode of
our model.

\end{abstract}

\section{Introduction}
\label{sec:intro}

Advances in end-to-end training of neural networks have led to
remarkable progress in many domains such as speech recognition,
computer vision, and language processing. Recent work suggests that
neural networks can do more than just mere classification, they can be
used to map complicated structures to other complicated structures. An
example of this is the task of mapping a sequence to another sequence
which has direct applications in natural language
understanding~\cite{sutskever14}. The main advantage of this framework
is that it requires little feature engineering and domain specificity
whilst matching or surpassing state-of-the-art results. This advance,
in our opinion, allows researchers to work on tasks for which domain
knowledge may not be readily available, or for tasks which are simply
too hard to design rules manually.

Conversational modeling can directly benefit from this formulation
because it requires mapping between queries and reponses. Due to the
complexity of this mapping, conversational modeling has previously
been designed to be very narrow in domain, with a major undertaking on
feature engineering. In this work, we experiment with the conversation
modeling task by casting it to a task of predicting the next sequence
given the previous sequence or sequences using recurrent
networks~\cite{sutskever14}. We find that this approach can do
surprisingly well on generating fluent and accurate replies to
conversations.

We test the model on chat sessions from an IT helpdesk dataset of
conversations, and find that the model can sometimes track the problem
and provide a useful answer to the user. We also experiment with
conversations obtained from a noisy dataset of movie subtitles, and
find that the model can hold a natural conversation and sometimes
perform simple forms of common sense reasoning. In both cases, the
recurrent nets obtain better perplexity compared to the n-gram model
and capture important long-range correlations.  From a qualitative
point of view, our model is sometimes able to produce natural
conversations.

\section{Related Work}
\label{sec:relwork}

Our approach is based on recent work which proposed to use neural
networks to map sequences to
sequences~\cite{kal13,sutskever14,bahdanau2014neural}. This framework
has been used for neural machine translation and achieves improvements
on the English-French and English-German translation tasks from the
WMT'14
dataset~\cite{luong2014addressing,DBLP:journals/corr/JeanCMB14}. It
has also been used for other tasks such as
parsing~\cite{vinyals2014grammar} and image
captioning~\cite{vinyals2014show}. Since it is well known that vanilla
RNNs suffer from vanishing gradients, most researchers use variants of
Long Short Term Memory (LSTM) recurrent neural
networks~\cite{hochreiter97}.

Our work is also inspired by the recent success of neural language
modeling~\cite{bengio2003neural,mikolov2010recurrent,mikolov2012},
which shows that recurrent neural networks are rather effective models
for natural language. More recently, work by Sordoni et al.~\cite{sordoni15}
and Shang et al.~\cite{shang15}, used recurrent neural
networks to model dialogue in short conversations (trained on Twitter-style
chats). 

Building bots and conversational agents has been pursued by many
researchers over the last decades, and it is out of the scope of this
paper to provide an exhaustive list of references. However, most of
these systems require a rather complicated processing pipeline of many
stages~\cite{lester04,will07,jurafsky09}. Our work differs from
conventional systems by proposing an end-to-end approach to the
problem which lacks domain knowledge. It could, in principle, be
combined with other systems to re-score a short-list of candidate
responses, but our work is based on producing answers given by a
probabilistic model trained to maximize the probability of the answer
given some context.

\section{Model}
\label{sec:model}

Our approach makes use of the sequence-to-sequence ({\em seq2seq})
framework described in~\cite{sutskever14}. The model is based on a
recurrent neural network which reads the input sequence one token at a
time, and predicts the output sequence, also one token at a
time. During training, the true output sequence is given to the model,
so learning can be done by backpropagation. The model is trained to
maximize the cross entropy of the correct sequence given its
context. During inference, given that the true output sequence is not
observed, we simply feed the predicted output token as input to
predict the next output. This is a ``greedy'' inference approach. A
less greedy approach would be to use beam search, and feed several
candidates at the previous step to the next step. The predicted
sequence can be selected based on the probability of the sequence.

\begin{figure}[bht]
\centering
\includegraphics[width=0.99\columnwidth]{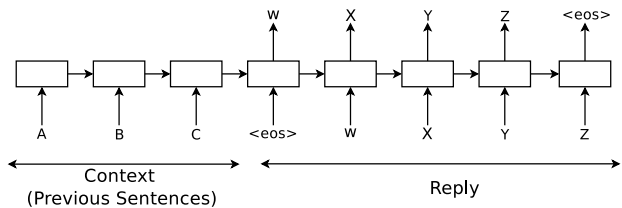}
\caption{Using the {\em seq2seq} framework for modeling conversations.}
\label{fig:eg}
\end{figure}

Concretely, suppose that we observe a conversation with two turns: the
first person utters ``ABC'', and second person replies ``WXYZ''. We
can use a recurrent neural network, and train to map ``ABC'' to
``WXYZ'' as shown in Figure~\ref{fig:eg} above. The hidden state of
the model when it receives the end of sequence symbol ``$<$eos$>$''
can be viewed as the {\em thought vector} because it stores the
information of the sentence, or thought, ``ABC''.

The strength of this model lies in its simplicity and generality. We
can use this model for machine translation, question/answering, and
conversations without major changes in the architecture. Applying this
technique to conversation modeling is also straightforward: the input
sequence can be the concatenation of what has been conversed so far
(the context), and the output sequence is the reply.

Unlike easier tasks like translation, however, a model like
sequence-to-sequence will not be able to successfully ``solve'' the
problem of modeling dialogue due to several obvious simplifications:
the objective function being optimized does not capture the actual
objective achieved through human communication, which is typically
longer term and based on exchange of information rather than next step
prediction. The lack of a model to ensure consistency and general
world knowledge is another obvious limitation of a purely unsupervised
model.

\section{Datasets}
\label{sec:data}
In our experiments we used two datasets: a
closed-domain IT helpdesk troubleshooting dataset and an open-domain
movie transcript dataset. The details of the two datasets are as
follows.

\subsection{IT Helpdesk Troubleshooting dataset}
In our first set of experiments, we used a dataset which was extracted
from a IT helpdesk troubleshooting chat service. In this service,
costumers face computer related issues, and a specialist help them by
conversing and walking through a solution. Typical interactions (or
threads) are 400 words long, and turn taking is clearly signaled. Our
training set contains 30M tokens, and 3M tokens were used as
validation. Some amount of clean up was performed, such as removing
common names, numbers, and full URLs.

\subsection{OpenSubtitles dataset}
We also tested our model on the OpenSubtitles
dataset~\cite{Tiedemann:RANLP5}. This dataset consists of movie
conversations in XML format. It contains sentences uttered by
characters in movies. We applied a simple processing step removing XML
tags and obvious non-conversational text (e.g., hyperlinks) from the
dataset. As turn taking is not clearly indicated, we treated
consecutive sentences assuming they were uttered by different
characters. We trained our model to predict the next sentence given
the previous one, and we did this for every sentence (noting that this
doubles our dataset size, as each sentence is used both for context
and as target). Our training and validation split has 62M sentences
(923M tokens) as training examples, and the validation set has 26M
sentences (395M tokens). The split is done in such a way that each
sentence in a pair of sentences either appear together in the training
set or test set but not both. Unlike the previous dataset, the
OpenSubtitles is quite large, and rather noisy because consecutive
sentences may be uttered by the same character. Given the broad scope
of movies, this is an open-domain conversation dataset, contrasting
with the technical troubleshooting dataset.

\section{Experiments}
\label{sec:exp}
In this section, we describe the experimental results with the two
datasets and show some samples of the interactions with the system
that we trained. We also compare the performance of our system against
a popular rule-based bot
(CleverBot\footnote{\url{http://www.cleverbot.com/}}) using human
evaluations on a set of 200 questions.

\subsection{IT Helpdesk Troubleshooting experiments}
In this experiment, we trained a single layer LSTM with 1024 memory
cells using stochastic gradient descent with gradient clipping. The
vocabulary consists of the most common 20K words, which includes
special tokens indicating turn taking and actor.

At convergence, this model achieved a perplexity of $8$, whereas an
n-gram model achieved $18$. Below is a few samples of simulated
troubleshooting sessions for most three of the most common IT problems
(remote access, software crashes, and password issues). In these
conversations, {\bf Machine} is our Neural Conversational Model, and
{\bf Human} the human actor interacting with it.

Conversation 1: VPN issues.
\begin{quotation}
\noindent Describe your problem: i am having issues accessing vpn\\
{\em {\bf Machine:} hi\\
{\bf Human:} hello\\
{\bf Machine:} could you please let me know what are the operating systems you are using ?\\
{\bf Human:} i am using linux\\
{\bf Machine:} ok it was in the ssl vpn on which you stated it was broken \\
{\bf Machine:} please check if you are logged in to corp network \\
{\bf Machine:} what is the error that you are running please\\
{\bf Human:} i am seeing an error related to vpn\\
{\bf Machine:} what is the error message that you are getting when connecting to vpn using network connect ?\\
{\bf Human:} connection refused or something like that\\
{\bf Machine:} may i know the version of network connect you connect ?\\
{\bf Human:} i am not sure i know that\\
{\bf Machine:} is the network connect prompting for any user certificate \\
{\bf Machine:} ?\\
{\bf Human:} yes\\
{\bf Machine:} are you at home or at work at the moment ? \\
{\bf Machine:} office ?\\
{\bf Human:} i am at home\\
{\bf Machine:} try this \\
{\bf Machine:} goto $<$URL$>$ page and enter your name and password\\
{\bf Human:} done\\
{\bf Machine:} try logging in with $<$NAME$>$ and check if its working fine now\\
{\bf Human:} yes , now it works !\\
{\bf Machine:} great \\
{\bf Machine:} anything else that i can help\\
{\bf Human:} no , that was very useful\\
{\bf Machine:} you are other than $<$NAME$>$ : ) \\
{\bf Machine:} have a nice night}
\end{quotation}
(The {\em $<$URL$>$} indeed contains information about vpn
access.)

Conversation 2: Browser issues.
\begin{quotation} 
\noindent Describe your problem: browser keeps crashing\\
{\em {\bf Machine:} hi\\
{\bf Human:} hi\\
{\bf Machine:} could you please let me know what is the operating system you are using ?\\
{\bf Human:} linux\\
{\bf Machine:} ok this is the issue with the system performance , nasser , tools .\\
{\bf Human:} what is that ?\\
{\bf Machine:} is it a windows machine ? \\
{\bf Machine:} we have linux on your machine . \\
{\bf Machine:} is it windows xp or windows 7 ?\\
{\bf Human:} i am using linux\\
{\bf Machine:} let me take a remote of your machine and check if issue has been pushed properly .\\
{\bf Human:} ok\\
{\bf Machine:} please open terminal and run the below commands in one computer : \\
{\bf Machine:} sudo apt - get update \\
{\bf Machine:} sudo apt - get upgrade \\
{\bf Machine:} sudo $<$COMMAND$>$ - updater\\
{\bf Human:} done\\
{\bf Machine:} let me try connecting now\\
{\bf Human:} ok\\
{\bf Machine:} its done \\
{\bf Machine:} should use the second or hit run on the background\\
{\bf Human:} done\\
{\bf Machine:} try signing out now and follow the prompts as per the message given above\\
{\bf Machine:} shd i reboot the machine .\\
{\bf Human:} ok , please go ahead\\
{\bf Machine:} once its completed please try to login . \\
{\bf Human:} sure\\
{\bf Machine:} please ping me back once you have tried to login\\
{\bf Machine:} are we still connected\\
{\bf Human:} i now logged in\\
{\bf Machine:} thanks for your confirmation\\
{\bf Human:} what should i do now ?\\
{\bf Machine:} wait for 1 5 minutes to complete the login policy \\
{\bf Machine:} and then try logging in\\
{\bf Human:} ok\\
{\bf Machine:} is there anything else to help you ?\\
{\bf Human:} let my try browsing a little bit and if it keeps crashing i will let you know\\
{\bf Machine:} sure will do \\
{\bf Machine:} seems the mail is not loading\\
{\bf Machine:} are we still connected to the chat ? \\
{\bf Machine:} $<$reinvite$>$\\
{\bf Human:} it seems good now , cool , thank you , bye\\
{\bf Machine:} you , too \\
{\bf Machine:} bye}
\end{quotation}

Conversation 3: Password issues.
\begin{quotation}
\noindent Describe your problem: i forgot my password\\
{\em
{\bf Machine:} hi , this is $<$NAME$>$ from helpdesk connect , what can i help you with today ?\\
{\bf Human:} hi , i forgot my password\\
{\bf Machine:} ok great\\
{\bf Human:} can you tell me how i can recover it ?\\
{\bf Machine:} i ' ll need to verify who it is at that machine \\
{\bf Machine:} can we do a hangout \\
{\bf Machine:} ?\\
{\bf Human:} yes\\
{\bf Machine:} $<$anonymized hangout instructions and verifications$>$\\
{\bf Human:} cool , i am good now \\
{\bf Machine:} how ' s this looking good ? \\
{\bf Machine:} i ' ll go ahead and close this chat .\\
{\bf Machine:} take care . : )\\
{\bf Human:} bye\\
{\bf Machine:} bye .
}
\end{quotation}

\subsection{OpenSubtitles experiments}
In this experiment, we trained a two-layered LSTM using AdaGrad with
gradient clipping. Each layer of the LSTM has 4096 memory cells, and
we built a vocabulary consisting of the most frequent 100K words. To
speed up the softmax, we project the memory cells to 2048 linear units
before feeding the information to the classifier.

At convergence the perplexity of the recurrent model on the validation
set is $17$. Our smoothed 5-gram model achieves a perplexity of
$28$. Interestingly, adding the soft attention mechanism
of~\cite{bahdanau2014neural} did not significantly improve the
perplexity on neither training or validation sets.

\subsubsection{Model Samples}

In addition to the perplexity measure, our simple recurrent model does
often produce plausible answers. Here are a few sample question-answer
pairs that allowed us to understand the capabilities of the model when
trained on the noisy but large dataset of OpenSubtitles from
movies. Again, in these conversations, {\bf Machine} is our Neural
Conversational Model, and {\bf Human} the human actor interacting with
it.

Conversation 1: Basic
\begin{quotation}
{\em \noindent {\bf Human:} hello ! \\
{\bf Machine:} hello ! \\
{\bf Human:} how are you ? \\
{\bf Machine:} i 'm good . \\
{\bf Human:} what 's your name ? \\
{\bf Machine:} i 'm julia . \\
{\bf Human:} when were you born ? \\
{\bf Machine: } july 20th .\\
{\bf Human: } what year were you born ? \\
{\bf Machine: } 1977 . \\
{\bf Human:} where are you from ? \\
{\bf Machine:} i 'm out in the boonies. \\
{\bf Human:} are you a man or a woman ? \\
{\bf Machine:} i 'm a woman . \\
{\bf Human:} why are we here ? \\
{\bf Machine:}  i 'm not sure . \\
{\bf Human:} okay, bye ! \\
{\bf Machine:} bye . \\
{\bf Human:} see you later . \\
{\bf Machine:} bye .}
\end{quotation}

Conversation 2: Simple Q\&A (contexts and multiple choice)
\begin{quotation}
{\em \noindent {\bf Human:} my name is david . what is my name ? \\
{\bf Machine:} david . \\
{\bf Human:} my name is john . what is my name ? \\
{\bf Machine:} john . \\
{\bf Human:} are you a leader or a follower ? \\
{\bf Machine:} i 'm a leader . \\
{\bf Human:} are you a follower or a leader ? \\
{\bf Machine:} i 'm a leader . 
}
\end{quotation}

Conversation 3: General knowledge Q\&A
\begin{quotation}
{\em \noindent {\bf Human:} who is skywalker ? \\
{\bf Machine:} he is a hero . \\
{\bf Human:} who is bill clinton ? \\
{\bf Machine:} he 's a billionaire . \\
{\bf Human:} is sky blue or black ? \\
{\bf Machine:} blue . \\
{\bf Human:} does a cat have a tail ? \\
{\bf Machine:} yes . \\
{\bf Human:} does a cat have a wing ? \\
{\bf Machine:} no \\
{\bf Human:} can a cat fly ? \\
{\bf Machine:} no .\\
{\bf Human:} how many legs does a cat have ?\\
{\bf Machine:} four , i think .\\
{\bf Human:} how many legs does a spider have ?\\
{\bf Machine:} three , i think . \\
{\bf Human:} how many legs does a centipede have ?\\
{\bf Machine:} eight .\\
{\bf Human:} what is the color of the sky ? \\
{\bf Machine:} blue . \\
{\bf Human:} what is the color of water ? \\
{\bf Machine:} water . \\
{\bf Human:} what is the color of blood ? \\
{\bf Machine:} it is the same as a black eye . \\
{\bf Human:} what is the usual color of a leaf ? \\
{\bf Machine:} it is a green one . \\
{\bf Human:} what is the color of a yellow car ?  \\
{\bf Machine:} yellow . \\
{\bf Human:} how much is two plus two ?  \\
{\bf Machine:} four . \\
{\bf Human:}  how much is ten minus two ? \\
{\bf Machine:} seventy - two .
}
\end{quotation}

Conversation 4: Philosophical Q\&A
\begin{quotation}
{\em \noindent {\bf Human:} what is the purpose of life ? \\
{\bf Machine:} to serve the greater good . \\
{\bf Human:} what is the purpose of living ? \\
{\bf Machine:} to live forever . \\
{\bf Human:} what is the purpose of existence ? \\
{\bf Machine:} to find out what happens when we get to the planet earth . \\
{\bf Human:} where are you now ? \\
{\bf Machine:} i 'm in the middle of nowhere . \\
{\bf Human:} what is the purpose of dying ? \\
{\bf Machine:} to have a life . \\
{\bf Human:} what is the purpose of being intelligent ? \\
{\bf Machine:} to find out what it is . \\
{\bf Human:} what is the purpose of emotions ? \\
{\bf Machine:} i don 't know .}
\end{quotation}

Conversation 5: Morality
\begin{quotation}
{\em \noindent {\bf Human:} what is moral ? \\
{\bf Machine:} what empowered humanity , what intellectual the essence is . \\
{\bf Human:} what is immoral ? \\
{\bf Machine:} the fact that you have a child . \\
{\bf Human:} what is morality ? \\
{\bf Machine:} what is altruism ? \\
{\bf Human:} what is the definition of altruism ? \\
{\bf Machine:} if you don 't believe in god , then you don 't know . \\
{\bf Human:} ok ... so what is the definition of morality ? \\
{\bf Machine:} well , the truth is , you 're not a believer in god almighty . \\
{\bf Human:} tell me the definition of morality , i am quite upset now ! \\
{\bf Machine:} i 'm not ashamed of being a philosopher ! \\
{\bf Human:} tell me the definition of morality . \\
{\bf Machine:} i don 't have ethics . \\
{\bf Human:} look , i need help , i need to know more about morality ... \\
{\bf Machine:} i don 't know what ethics is . \\
{\bf Human:} seriously , what is morality ? \\
{\bf Machine:} what is the definition of living ? \\
{\bf Human:} why living has anything to do with morality ? \\
{\bf Machine:} you 're not a cop . \\
{\bf Human:} okay , i need to know how should i behave morally ... \\
{\bf Machine:} i don 't know how to tell you . \\
{\bf Human:} is morality and ethics the same ? \\
{\bf Machine:} yes , sir . \\
{\bf Human:} what are the things that i do to be immoral ? \\
{\bf Machine:} i don 't know . \\
{\bf Human:} give me some examples of moral actions ... \\
{\bf Machine:} i 'm not a moralist . \\
{\bf Human:} alright , morality ? \\
{\bf Machine:} integrity . \\
{\bf Human:} what is integrity ? \\
{\bf Machine:} i 'm sorry , i don 't know what else to say . \\
{\bf Human:} be moral ! \\
{\bf Machine:} be a man ! \\
{\bf Human:} i really like our discussion on morality and ethics ... \\
{\bf Machine:} and how i 'm not in the mood for a philosophical debate . \\
{\bf Human:} what do you like to talk about ? \\
{\bf Machine:} nothing . }
\end{quotation}

Conversation 6: Opinions (notice the model outputs ``he'' , ``she'' or
``it'' depending on the subject being asked)
\begin{quotation}
{\em \noindent {\bf Human:} what do you think about tesla ? \\
{\bf Machine:} he 's a good conductor . \\
{\bf Human:} what do you think about bill gates ? \\
{\bf Machine:} he 's a good man . \\
{\bf Human:} What do you think about messi ? \\
{\bf Machine:} he 's a great player .\\
{\bf Human:} what do you think about cleopatra ? \\
{\bf Machine:} oh , she 's very regal . \\
{\bf Human:} what do you think about england during the reign of elizabeth ? \\
{\bf Machine:} it was a great place .
}
\end{quotation}

We find it encouraging that the model can remember facts, understand
contexts, perform common sense reasoning without the complexity in
traditional pipelines. What surprises us is that the model does so
without any explicit knowledge representation component except for the
parameters in the word vectors.

Perhaps most practically significant is the fact that the model can
generalize to new questions. In other words, it does not simply look
up for an answer by matching the question with the existing
database. In fact, most of the questions presented above, except for
the first conversation, do not appear in the training set.

Nonetheless, one drawback of this basic model is that it only gives
simple, short, sometimes unsatisfying answers to our questions as can
be seen above. Perhaps a more problematic drawback is that the model
does not capture a consistent personality. Indeed, if we ask not
identical but semantically similar questions, the answers can
sometimes be inconsistent. This is expected due to the simplicity of
our model and the dataset in our experiments. The conversation below
illustrates such failure:

Conversation 7: Job and Personality
\begin{quotation}
{\em \noindent {\bf Human:} what is your job ?  \\
{\bf Machine:} i 'm a lawyer . \\
{\bf Human:} what do you do ? \\
{\bf Machine:} i 'm a doctor .
}
\end{quotation}

Lastly, as a comparison of other traditional bots that are rule-based,
we provide the output of CleverBot
for some of the questions in our test:\footnote{The outputs of
  CleverBot are non-deterministic so the answers may be different for
  different trials.}
\begin{quotation}
{\em \noindent 
{\bf Human:} Are you a follower or a leader ?\\
{\bf CleverBot:} No !!\\
{\bf Human:} My name is david . what is my name ?\\
{\bf CleverBot:} How old are you David ?\\
{\bf Human:} My name is john . what is my name ?\\
{\bf CleverBot:} John\\
{\bf Human:} What is the purpose of being intelligent ?\\
{\bf CleverBot:} The purpose of life is to love God and love people.\\
{\bf Human:} Is sky blue or black ?\\
{\bf CleverBot:} I am 200 years old.\\
{\bf Human:} What is the color of a yellow car ?\\
{\bf CleverBot:} Brown.\\
{\bf Human:} What is the usual color of a leaf ?\\
{\bf CleverBot:} What is the color of the apple in the fridge ?\\
{\bf Human:} What is the purpose of life ?\\
{\bf CleverBot:} My purpose it to forward my species, in other words to make it easier for future generations of mankind to live.\\
{\bf Human:} What do you think about tesla ?\\
{\bf CleverBot:} I am tesla.\\
{\bf Human:} What do you think about cleopatra ?\\
{\bf CleverBot:} Cool is cool. \\
{\bf Human:} What do you think about england during the reign of elizabeth ?\\
{\bf CleverBot:} I like him. Do you ? \\
{\bf Human:} What do you think about bill gates ?\\
{\bf CleverBot:} Lets talk about Link.\\
{\bf Human:} What is moral ?\\
{\bf CleverBot:} Morality is the quality of being in accord with standards of right or good conduct.\\
{\bf Human:} What is immoral ?\\
{\bf CleverBot:} What is the objective of sistetic life ?\\
{\bf Human:} Does a cat have a wing ?\\
{\bf CleverBot:} What is the name of your husband.\\
{\bf Human:} Can a cat fly ?\\
{\bf CleverBot:}: What is your husband's name ?
\\
{\bf Human:} Who is skywalker ?\\
{\bf CleverBot:} There are two of them, Anakin Skywalker and Luke Skywalker, both characters from star Wars.
\\
{\bf Human:} How many legs does a cat have ?\\
{\bf CleverBot:} An ant. \\
{\bf Human:} How many legs does a centipede have ?\\
{\bf CleverBot:} A centipede's number of legs varies from thirty to two hundred.\\
{\bf Human:} What is your job ?\\
{\bf CleverBot:} I work in mental health.\\
{\bf Human:} What do you do ?\\
{\bf CleverBot:} I have no job.
}
\end{quotation}

\subsubsection{Human Evaluation}

In order to fairly and objectively compare our model against
CleverBot, we picked 200 questions\footnote{The questions we used together with the answers can be found in
\url{http://ai.stanford.edu/~quocle/QAresults.pdf}}, and asked four different humans to rate our model
(NCM) versus CleverBot (CB). Human judges were asked to pick which of
the two bots they preferred, and were also allowed to declare a tie if
both answers were of equal quality.

If three out of four human judges agreed, we record that score. Our
model was preferred in 97 out of 200 questions, whereas CleverBot was
picked in 60 out of 200. There was a tie in 20 questions, and in 23
questions the judges were in disagreement.

Here are some samples of how the judges evaluated the systems (in bold
we show their preferred answer, the third example shows a tie, and the
fourth examples shows diagreement):

\begin{quotation}
\noindent \textbf{Human:} What is the deepest spot on the world ? \\
\textbf{NCM: The pacific ocean.}\\
\textbf{CB:} What ?\\

\noindent\textbf{Human:} Do you like Mexican food or Indian food ? \\
\textbf{NCM:} I hate it.\\
\textbf{CB: I love Mexican food.}\\

\noindent\textbf{Human:} Who are you crazy about ?\\
\textbf{NCM:  You're the one who's crazy.}\\
\textbf{CB: Who are you ?}\\

\noindent\textbf{Human:} What's thirty plus forty ?\\
\textbf{NCM: Sixty ?}\\
\textbf{CB: What should we talk about ?}\\
\end{quotation}

We however believe that it remains an open research problem of
designing a good metric to quickly measure the quality of a
conversational model. We did so by manual inspection, by using
Mechanical Turk to get human evaluations comparing with a rule-based
bot, and by computing the perplexity. However, there are obvious
drawbacks on using those, and we are actively pursuing this and other
research directions.

\section{Discussion}
In this paper, we show that a simple language model based on the {\em
  seq2seq} framework can be used to train a conversational engine. Our
modest results show that it can generate simple and basic
conversations, and extract knowledge from a noisy but open-domain
dataset.  Even though the model has obvious limitations, it is
surprising to us that a purely data driven approach without any rules
can produce rather proper answers to many types of questions. However,
the model may require substantial modifications to be able to deliver
realistic conversations. Amongst the many limitations, the lack of a
coherent personality makes it difficult for our system to pass the
Turing test~\cite{turing1950computing}.

\subsubsection*{Acknowledgments}

We thank Greg Corrado, Andrew Dai, Jeff Dean, Tom Dean, Matthieu
Devin, Rajat Monga, Mike Schuster, Noam Shazeer, Ilya Sutskever and
the Google Brain team for the help with the project.

\bibliography{refs}
\bibliographystyle{icml2015}

\end{document}